\documentclass[runningheads]{llncs}
\usepackage[T1]{fontenc}
\usepackage{graphicx}
\usepackage{color}
\usepackage{amsmath}
\usepackage{mathtools}
\usepackage{amsfonts}
\usepackage[labelfont=bf]{caption}
\usepackage[labelformat=simple,labelfont=bf]{subcaption}

\usepackage{booktabs,multirow}
\usepackage{array}
\usepackage{pifont}
\usepackage{bbding}
\usepackage[pagebackref=true,breaklinks=true,colorlinks=true,bookmarks=false,citecolor=blue,urlcolor=blue,linkcolor=blue]{hyperref}
\usepackage[misc]{ifsym}

\makeatletter
\newcommand*{\rom}[1]{\expandafter\@slowromancap\romannumeral #1@}
\makeatother

\begin{document}
%
\title{Neural Rendering for Stereo 3D Reconstruction of Deformable Tissues in Robotic Surgery}
\titlerunning{Neural Rendering for Stereo 3D Reconstruction in Robotic Surgery}
%
\author{Yuehao Wang\inst{1} \and Yonghao Long\inst{1} \and Siu Hin Fan\inst{2} \and Qi Dou\inst{1}\textsuperscript{(\Letter)}}

\authorrunning{Y. Wang et al.}
%
\institute{}
\institute{Dept. of Computer Science and Engineering, The Chinese University of Hong Kong \\ 
\and Dept. of Biomedical Engineering, The Chinese University of Hong Kong\\}
\maketitle              
\begin{abstract}
Reconstruction of the soft tissues in robotic surgery from endoscopic stereo videos is important for many applications such as intra-operative navigation and image-guided robotic surgery automation. Previous works on this task mainly rely on SLAM-based approaches, which struggle to handle complex surgical scenes. Inspired by recent progress in neural rendering, we present a novel framework for deformable tissue reconstruction from binocular captures in robotic surgery under the single-viewpoint setting. Our framework adopts dynamic neural radiance fields to represent deformable surgical scenes in MLPs and optimize shapes and deformations in a learning-based manner. In addition to non-rigid deformations, tool occlusion and poor 3D clues from a single viewpoint are also  particular challenges in soft tissue reconstruction. To overcome these difficulties, we present a series of strategies of tool mask-guided ray casting, stereo depth-cueing ray marching and stereo depth-supervised optimization.
With experiments on DaVinci robotic surgery videos, our method significantly outperforms the current state-of-the-art reconstruction method for handling various complex non-rigid deformations.
To our best knowledge, this is the first work leveraging neural rendering for surgical scene 3D reconstruction with remarkable potential demonstrated. Code is available at: \url{https://github.com/med-air/EndoNeRF}.

\end{abstract}

\section{Introduction}

Surgical scene reconstruction from endoscope stereo video is an important but difficult task in robotic minimally invasive surgery. 
It is a prerequisite for many downstream clinical applications, including intra-operative navigation and augmented reality, 
surgical environment simulation, immersive education, and robotic surgery automation~\cite{chen2018slam,lu2021super,penza2017envisors,tang2018augmented}.
Despite much recent progress~\cite{liu2020reconstructing,recasens2021endo,tukra2021see,wei2021laparoscopic,wei2021stereo,zhou2021real}, several key challenges still remain unsolved.
First, surgical scenes are deformable with significant topology changes, requiring dynamic reconstruction to capture a high degree of non-rigidity.
Second, endoscopic videos show sparse viewpoints due to constrained camera movement in confined space, resulting in limited 3D clues of soft tissues.
Third, the surgical instruments always occlude part of the soft tissues, which affects the completeness of surgical scene reconstruction.

Previous works~\cite{brandao2021hapnet,luo2022unsupervised} explored the effectiveness of surgical scene reconstruction via depth estimation. Since most of the endoscopes are equipped with stereo cameras, depth can be estimated from binocular vision. Follow-up SLAM-based methods \cite{song2017dynamic,zhou2019real,zhou2021emdq} fuse depth maps in 3D space to reconstruct surgical scenes under more complex settings. Nevertheless, these methods either hypothesize scenes as static or surgical tools not present, limiting their practical use in real scenarios.
Recent work SuPer \cite{li2020super} and E-DSSR \cite{long2021dssr} present frameworks consisting of tool masking, stereo depth estimation and SurfelWarp \cite{gao2019surfelwarp} to perform single-view 3D reconstruction of deformable tissues. However, all these methods track deformation based on a sparse warp field \cite{newcombe2015dynamicfusion}, which degenerates when deformations are significantly beyond the scope of non-topological changes.


As an emerging technology, neural rendering \cite{kato2018neural,tewari2020state,tewari2021advances} is recently developed to break through the limited performance of traditional 3D reconstruction by leveraging differentiable rendering and neural networks. In particular, neural radiance fields (NeRF) \cite{mildenhall2020nerf}, a popular pioneering work of neural rendering, proposes to use \textit{neural implicit field} for continuous scene representations and achieves great success in producing high-quality view synthesis and 3D reconstruction on diverse scenarios \cite{martin2021nerf,mildenhall2020nerf,niemeyer2021giraffe}. Meanwhile, recent variants of NeRF \cite{park2021nerfies,park2021hypernerf,pumarola2021d} targeting dynamic scenes have managed to track deformations through various neural representations on non-rigid objects.

In this paper, we endeavor to reconstruct highly deformable surgical scenes captured from single-viewpoint stereo endoscopes. We embark on adapting the emerging neural rendering framework to the regime of deformable surgical scene reconstruction. We summarize our contributions as follows:  1) To accommodate a wide range of geometry and deformation representations on soft tissues, we leverage neural implicit fields to represent dynamic surgical scenes. 2) To address the particular tool occlusion problem in surgical scenes, we design a new mask-guided ray casting strategy for resolving tool occlusion. 3) We incorporate a depth-cueing ray marching and depth-supervised optimization scheme, using stereo prior to enable neural implicit field reconstruction for single-viewpoint input. To the best of our knowledge, this is the first work introducing cutting-edge neural rendering to surgical scene reconstruction. We evaluate our method on 6 typical in-vivo surgical scenes of robotic prostatectomy. Compared with previous methods, our results exhibit great performance gain, both quantitatively and qualitatively, on 3D reconstruction and deformation tracking of surgical scenes.

\section{Method}

\subsection{Overview of the Neural Rendering-based Framework}

\begin{figure}[t!]
    \centering
    \includegraphics[width=0.9\textwidth]{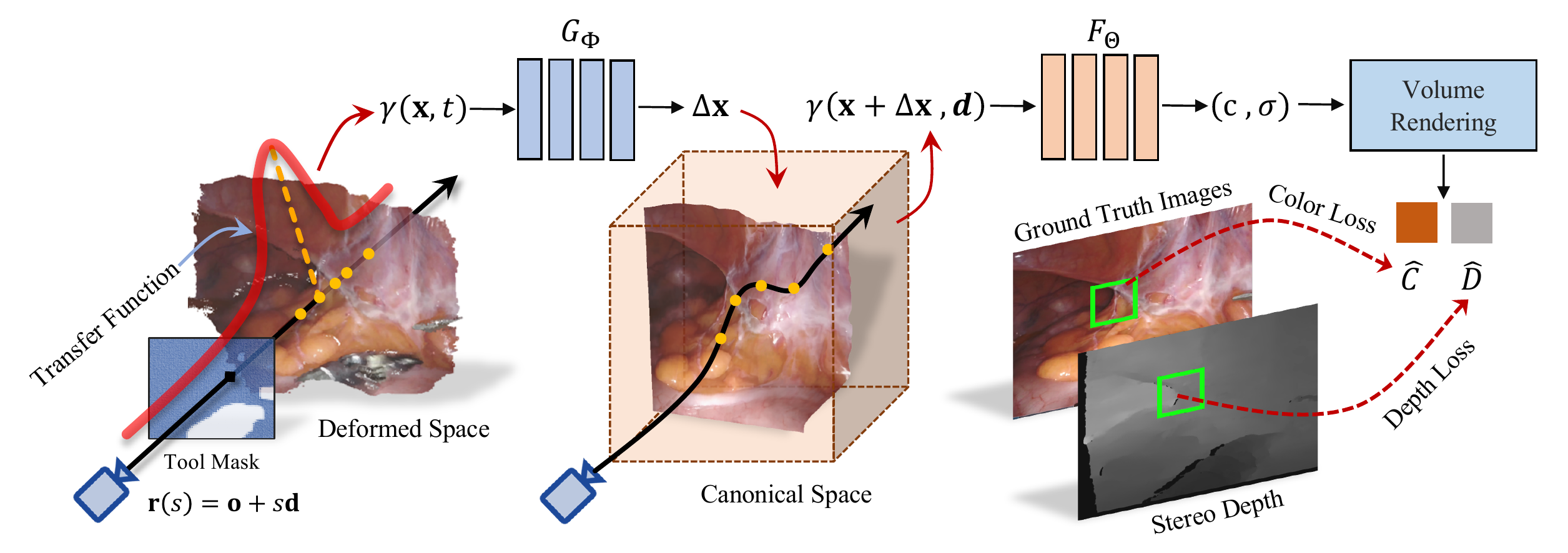}
    \caption{Illustration of our proposed novel approach of neural rendering for stereo 3D reconstruction of deformable tissues in robotic surgery.}
    \label{fig:pipeline}
\end{figure}

Given a single-viewpoint stereo video of a dynamic surgical scene, we aim to reconstruct 3D structures and textures of surgical scenes without occlusion of surgical instruments. We denote $\{(\boldsymbol{I}^l_i, \boldsymbol{I}^r_i)\}_{i=1}^T$ as a sequence of input stereo video frames, where $T$ is the total number of frames and $(\boldsymbol{I}^l_i, \boldsymbol{I}^r_i)$ is the pair of left and right images at the $i$-th frame. The video duration is normalized to $[0, 1]$. Thus, time of the $i$-th frame is $i/T$. We also extract binary tool masks $\{\boldsymbol{M}_i\}_{i=1}^T$ for the left views to identify the region of surgical instruments. To utilize stereo clues, we estimate coarse depth maps $\{\boldsymbol{D}_i \}_{i=1}^T$ for the left views from the binocular captures. We follow the modeling in D-NeRF \cite{pumarola2021d} and represent deformable surgical scenes as a canonical neural radiance field along with a time-dependent neural displacement field (cf. Sec.~\ref{sec:scene_representation}).
In our pipeline, each training iteration consists of the following six stages: \romannumeral 1) randomly pick a frame for training, \romannumeral 2) run tool-guided ray casting (cf. Sec.~\ref{sec:ray_casting}) to shoot camera rays into the scene, \romannumeral 3) sample points along each camera ray via depth-cueing ray marching (cf. Sec.~\ref{sec:ray_marching}), \romannumeral 4) send sampled points to networks to obtain color and space occupancy of each point, \romannumeral 5) evaluate volume rendering integral on sampled points to produce rendering results, \romannumeral 6) optimize the rendering loss plus depth loss to reconstruct shapes, colors and deformations of the surgical scene (cf. Sec.~\ref{sec:optimization}). The overview of key components in our approach is illustrated in Fig. \ref{fig:pipeline}. We will describe the detailed methods in the following subsections.

\subsection{Deformable Surgical Scene Representations}
\label{sec:scene_representation}

We represent a surgical scene as a canonical radiance field and a time-dependent displacement field. Accordingly, each frame of the surgical scene can be regarded as a deformation of the canonical field. The canonical field, denoted as $F_\Theta(\mathbf{x}, \mathbf{d})$, is an 8-layer MLP with network parameter $\Theta$, mapping coordinates $\mathbf{x}\in \mathbb{R}^3$ and unit view-in directions $\mathbf{d}\in \mathbb{R}^3$ to RGB colors $c(\mathbf{x}, \mathbf{d})\in\mathbb{R}^3$ and space occupancy $\sigma(\mathbf{x})\in\mathbb{R}$.
The time-dependent displacement field $G_\Phi(\mathbf{x}, t)$ is encoded in another 8-layer MLP with network parameters $\Phi$ and maps input space-time coordinates $(\mathbf{x}, t)$ into displacement between the point $\mathbf{x}$ at time $t$ and the corresponding point in the canonical field.
For any time $t$, the color and occupancy at $\mathbf{x}$ can be retrieved as $F_\Theta(\mathbf{x} + G_\Phi(\mathbf{x}, t), \mathbf{d})$. Compared with other dynamic modeling approaches \cite{park2021nerfies,park2021hypernerf}, a displacement field is sufficient  to explicitly and physically express all tissue deformations.
To capture high-frequency details, we use positional encoding $\gamma(\cdot)$ to map the input coordinates and time into Fourier features \cite{tancik2020fourier} before  feeding them to the networks.




\subsection{Tool Mask-Guided Ray Casting}
\label{sec:ray_casting}


With scene representations, we further leverage the differentiable volume rendering used in NeRF to yield renderings for supervision. The differentiable volume rendering begins with shooting a batch of camera rays into the surgical scene from a fixed viewpoint at an arbitrary time $t$. Every ray is formulated as $\mathbf{r}(s) = \mathbf{o} + s \mathbf{d}$, where $\mathbf{o}$ is a fixed origin of the ray, $\mathbf{d}$ is the pointing direction of the ray and $s$ is the ray parameter. In the original NeRF, rays are shot towards a batch of randomly selected pixels on the entire image plane. However, there are many pixels of surgical tools on the captured images, while our goal is to reconstruct underlying tissues. Thus, training on these tool pixels is unexpected.
Our main idea for solving this issue is to bypass those rays traveling through tool pixels over the training stage. We utilize binary tool masks $\{\boldsymbol{M}_i\}_{i=1}^T$, where 0 stands for tissue pixels and 1 stands for tool pixels, to inform which rays should be neglected. In this regard, we create importance maps $\{\boldsymbol{\mathcal{V}}\}_{i=1}^T$ according to $\boldsymbol{M}_i$ and perform importance sampling to avoid shooting rays for those pixels of surgical tools. Eq. (\ref{eqn:init_importance_maps}) exhibits the construction of importance maps, where $\otimes$ is element-wise multiplication, $\lVert \cdot \rVert_F$ is Frobenius norm and $\mathbf{1}$ is a matrix with the same shape as $\boldsymbol{M}_i$ while filled with ones:
\begin{align}
\begin{split}
    \boldsymbol{\mathcal{V}}_i = \boldsymbol{\Lambda} \otimes \big(\mathbf{1} - \boldsymbol{M}_i\big), \quad \boldsymbol{\Lambda} = \bigg(\mathbf{1} + \sum\nolimits_{j=1}^{T}\boldsymbol{M}_j \bigg/ \big\lVert \sum\nolimits_{j=1}^{T}\boldsymbol{M}_j\big\rVert_F\bigg).
    \label{eqn:init_importance_maps}
\end{split}
\end{align}
The $\mathbf{1} - \boldsymbol{M}_i$ term initializes the importance of tissue pixels to 1 and the importance of tool pixels to 0. To balance the sampling rate of occluded pixels across frames, the scaling term $\boldsymbol{\Lambda}$ specifies higher importance scaling for those tissue areas with higher occlusion frequencies.
Normalizing each importance map as $\widehat{\boldsymbol{\mathcal{V}}}_i = \boldsymbol{\mathcal{V}}_i / \lVert\boldsymbol{\mathcal{V}}_i \rVert_F$ will yield a probability mass function over the image plane. During our ray casting stage for the $i$-th frame, we sample pixels from the distribution $\widehat{\boldsymbol{\mathcal{V}}}_i$ using inverse transform sampling and cast rays towards these sampled pixels. In this way, the probability of shooting rays for tool pixels is guaranteed to be zero as the importance of tool pixels is constantly zero.

\subsection{Stereo Depth-Cueing Ray Marching}
\label{sec:ray_marching}

After shooting camera rays over tool occlusion, we proceed ray marching to sample points in the space. Specifically, we discretize each camera ray $\mathbf{r}(s)$ into batch of points $\{\mathbf{x}_j \lvert \mathbf{x}_j = \mathbf{r}(s_j) \}_{j=1}^{m}$ by sampling a sequence of ray steps $s_1 \leq s_2 \leq \dots \leq s_m$. The original NeRF proposes hierarchical stratified sampling to obtain $\{s_j\}_{j=1}^{m}$. However, this sampling strategy hardly exploits accurate 3D structures when NeRF models are trained on single-view input. Drawing inspiration from early work in iso-surface rendering \cite{kniss2003gaussian}, we create Gaussian transfer functions with stereo depth to guide point sampling near tissue surfaces. For the $i$-th frame,  the transfer function for a ray $\mathbf{r}(s)$ shooting towards pixel $(u, v)$ is formulated as:
\begin{align}
\begin{split}
    \delta(s; u, v, i) = \exp\big({-(s - \boldsymbol{D}_i[u, v])^2 \big/ 2\xi^2}\big).
    \label{eqn:stereo_depth_transfer_func}
\end{split}
\end{align}
The transfer function $\delta(s; u, v, i)$ depicts an impulse distribution that continuously allocates sampling weights for every location on $\mathbf{r}(s)$. The impulse is centered at $\boldsymbol{D}_i[u, v]$, i.e., the depth at the $(u,v)$ pixel. The width of the impulse is controlled by the hyperparameter $\xi$, which is set to a small value to mimic Dirac delta impulse. In our ray marching, $s_1 \leq s_2, \leq \dots \leq s_m$ are drawn from the normalized impulse distribution $\frac{1}{\xi\sqrt{2\pi}}\delta(s; u, v, i)$. By this means, sampled points are concentrated around tissue surfaces, imposing stereo prior in rendering.



\subsection{Optimization for Deformable Radiance Fields}
\label{sec:optimization}


Once we obtain the sampled points in the space, the emitted color $\widehat{C}$ and optical depth $\widehat{D}$ of a camera ray $\mathbf{r}(s)$ can be evaluated by volume rendering \cite{kajiya1984ray} as:
\begin{align}
\begin{split}
    &\widehat{C}\big(\mathbf{r}(s)\big) = \sum\nolimits_{j=1}^{m-1} w_jc(\mathbf{x}_j, \mathbf{d}),
    \quad \widehat{D}\big(\mathbf{r}(s)\big) = \sum\nolimits_{j=1}^{m-1} w_js_j ,\\
    &w_j = (1-\exp{(-\sigma(\mathbf{x}_j)}\Delta s_j))\exp{(-\sum\nolimits_{k=1}^{j-1}\sigma(\mathbf{x}_k)\Delta s_k)},~\Delta s_j = s_{j+1} - s_j .
    \label{eqn:volrend_color_int}
\end{split}
\end{align}
To reconstruct the canonical and displacement fields from single-view captures, we optimize the network parameters $\Theta$ and $\Phi$ by jointly supervising the rendered color and optical depth \cite{deng2021depth}.
Specifically, the loss function for training the networks is defined as:
\begin{align}
\begin{split}
    \mathcal{L}(\mathbf{r}(s)) = \big\lVert \widehat{C}(\mathbf{r}(s)) - \boldsymbol{I}_i[u, v] \big\rVert_2^2 + \lambda \big\lvert \widehat{D}(\mathbf{r}(s)) - \boldsymbol{D}_i[u, v] \big\rvert ,
    \label{eqn:loss_func}
\end{split}
\end{align}
where $(u,v)$ is the location of the pixel that $\mathbf{r}(s)$ shoots towards, $\lambda$ is a hyperparameter weighting the depth loss.

  

Last but not least, we conduct statistical depth refinement to handle corrupt stereo depth caused by fuzzy pixels and specular highlights on the images of surgical scenes.
Direct supervision on the estimated depth will overfit corrupt depth in the end, leading to abrupt artifacts in reconstruction results (Fig. \ref{fig:ablation_depth_modules}). Our preliminary findings reveal that our model at the early training stage would produce smoother results both in color and depth since the underfitting model tends to average learned colors and occupancy. Thus, minority corrupt depth is smoothed by majority normal depth. Based on this observation, we propose to patch the corrupt depth with the output from underfitting radiance fields. Denoting $\widehat{\boldsymbol{D}}^K_i$ as the underfitting output depth maps for the $i$-th frame after $K$ iterations of training, we firstly find residual maps through $\boldsymbol{\epsilon}_i = \lvert \widehat{\boldsymbol{D}}_i^K - \boldsymbol{D}_i \rvert$, then we compute a probabilistic distribution over the residual maps. After that, we set a small number $\alpha \in [0, 1]$ and locate those pixels with the last $\alpha$-quantile residuals. Since those located pixels statistically correspond to large residuals, we can identify them as occurrences of corrupt depth. Finally, we replace those identified corrupt depth pixels with smoother depth pixels in $\widehat{\boldsymbol{D}}^K_i$. After this refinement procedure, the radiance fields are optimized on the patched depth maps in the subsequent training iterations, alleviating corrupt depth fitting.




\section{Experiments}

\subsection{Dataset and Evaluation Metrics}

We evaluate our proposed method on typical robotic surgery stereo videos from 6 cases of our in-house DaVinci robotic prostatectomy data. We totally extracted 6 clips with a total of 807 frames. Each clip lasts for $4 \! \sim \! 8s$ with $15fps$. Each case is captured from stereo cameras at a single viewpoint and encompasses challenging scenes with non-rigid deformation and tool occlusion. Among the selected 6 cases, 2 cases contain traction on thin structures such as fascia, 2 cases contain significant pushing and pulling of tissue, and 2 cases contain tissue cutting, which altogether present the typical soft tissue situations in robotic surgery.
For comparison, we take the most recent state-of-the-art surgical scene reconstruction method of E-DSSR \cite{long2021dssr} as a strong comparison. 
For qualitative evaluation,
We exhibit our reconstructed point clouds and compare textural and geometric details obtained by different methods. We also conduct an ablation study on our depth-related modules through qualitative comparison. 
Due to clinical regulation in practice, it is impossible to collect ground truth depth for numerical evaluation on 3D structures. Following the evaluation method in~\cite{long2021dssr} and wide literature in neural rendering, we alternatively use photometric errors, including PSNR, SSIM and LPIPS, as evaluation metrics for quantitative comparisons.




\subsection{Implementation Details}
In our implementation, we empirically set the width of the transfer function $\xi \!= \!1$, the weight of depth loss $\lambda \!=\!1$, depth refinement iteration $K\!=\!4000$ and $\alpha\!=\!0.1$. Other training hyper-parameters follow the settings in the state-of-the-art D-NeRF~\cite{pumarola2021d}. We calibrate the endoscope in advance to acquire its intrinsics. In all of our experiments, tool masks are obtained by manually labeling and coarse stereo depth maps are generated by STTR-light \cite{li2021revisiting} pretrained on Scene Flow. We optimize each model over $100K$ iterations on a single case. To recover explicit geometry from implicit fields, we render optimized radiance fields to RGBD maps, smooth rendered depth maps via bilateral filtering, and back-project RGBD into point clouds based on the endoscope intrinsics. 

\subsection{Qualitative and Quantitative Results}


\begin{figure}[t!]
    \centering
    \begin{subfigure}[a]{\textwidth}
         \centering
         \includegraphics[width=0.95\textwidth]{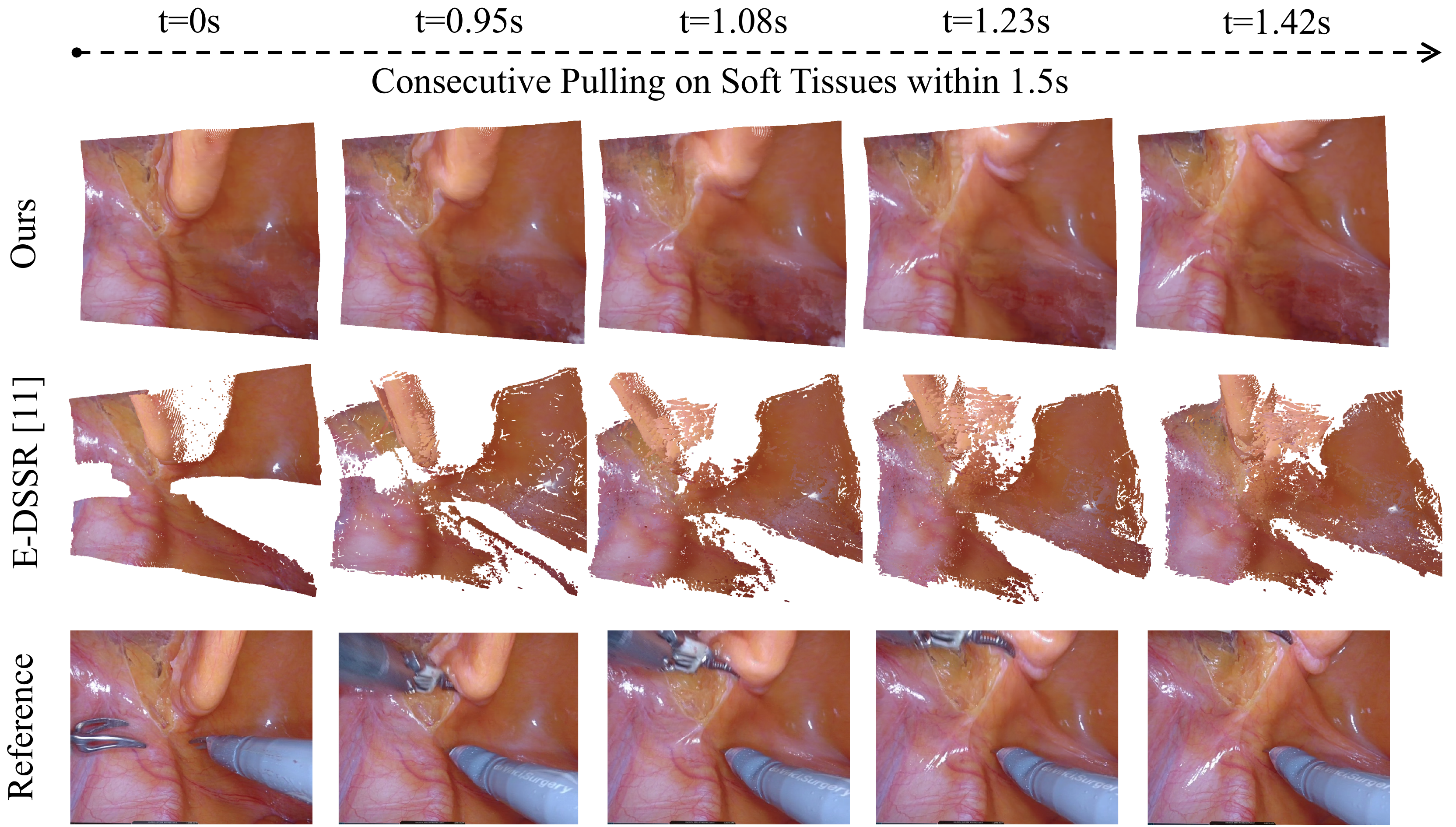}
         \caption{Results on the case ``pulling tissues'', where soft tissues are drastically pulled within 2s. We exhibit 5 reconstruction results of our method and E-DSSR over time.}
         \label{fig:pulling_tissues}
    \end{subfigure}
    \begin{subfigure}[b]{\textwidth}
         \centering
         \includegraphics[width=0.95\textwidth]{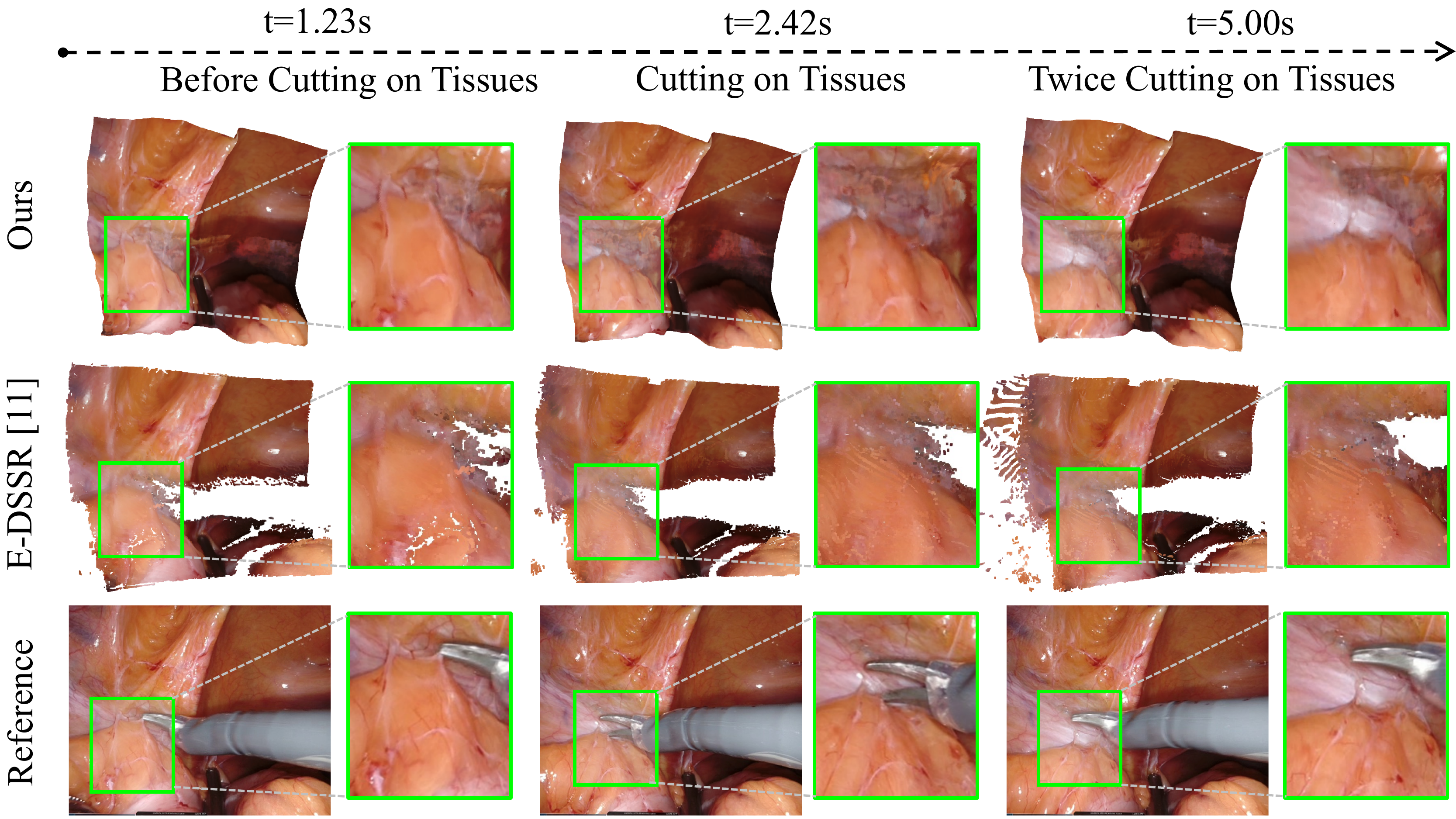}
         \caption{Results on the case ``cutting tissues twice''. We show 3 frames corresponding to no deformation, cutting once and cutting twice, respectively. The close-ups of the cutting areas display reconstructed tissue details before and after cutting.}
         \label{fig:twice_cutting_tissues}
    \end{subfigure}
    \caption{Qualitative comparisons of 2 cases, demonstrating reconstruction of soft tissues with large deformations and topology changes.}
    \label{fig:quali_topology_change}
    \label{fig:result} 
\end{figure}

For qualitative evaluation, Fig.~\ref{fig:result} illustrates the reconstruction results of our approach and the comparison method, along with a reference to the original video.
In the test case of Fig. \ref{fig:pulling_tissues}, the tissues are pulled by surgical instruments, yielding relatively large deformations. Benefitting from the underlying continuous scene representations, our method can reconstruct water-tight tissues without being affected by the tool occlusion. More importantly, per-frame deformations are captured continuously, achieving stable results over the episode of consecutive pulling. In contrast, the current state-of-the-art method~\cite{long2021dssr} could not fully track these large deformations and its reconstruction results include holes and noisy points under such a challenging situation. We further demonstrate a more difficult case in Fig. \ref{fig:twice_cutting_tissues} which includes soft tissue cutting with topology changes. From the reconstruction results, it is observed that our method manages to track the detailed cutting procedures, owing to the powerful neural representation of displacement fields. In addition, it can bypass the issue of tool occlusion and recover the hidden tissues, which is cooperatively achieved by our mask-guided ray casting and the interpolation property of neural implicit fields. 
On the other hand, the comparison method is not able to capture these small changes on soft tissues nor patch all the tool-occluded areas.
Table~\ref{tab:quanti_eval} summarizes our quantitative experiments, showing overall performance on the dataset. Our method dramatically outperforms E-DSSR by $\uparrow$~16.433 PSNR, $\uparrow$~0.295 SSIM and $\downarrow$~0.342 LPIPS. To assess the contribution of the dynamics modeling, we also evaluate our model without neural displacement field (Ours w/o D). As expected, removing this component leads to a noticeable performance drop, which reflects the effectiveness of the displacement modeling.

\begin{figure}[t!]
    \centering
    \includegraphics[width=0.95\textwidth]{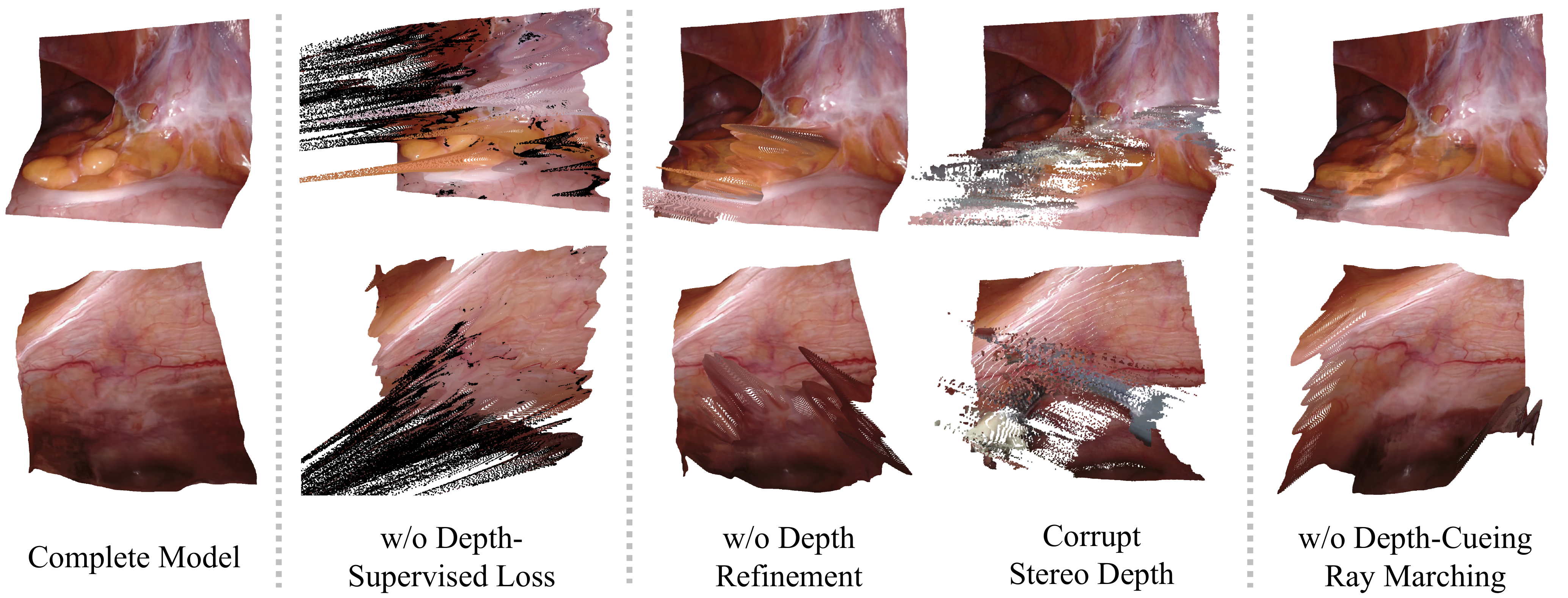}
    \caption{Ablation study on our depth-related modules, i.e., depth-supervised loss, depth refinement and depth-cueing ray marching.}
    \label{fig:ablation_depth_modules}
\end{figure}

\begin{table}[t!]
    \centering
    \caption{Quantitative evaluation on photometric errors of the dynamic reconstruction on metrics of PSNR, SSIM and LPIPS.}
    \begin{tabular}{ >{\raggedright\arraybackslash}m {6.5em} | >{\centering\arraybackslash}m {9.2em} | >{\centering\arraybackslash}m {9.2em} |  >{\centering\arraybackslash}m {9.2em} }
    \toprule[0.5pt]
        \textbf{Methods} & \textbf{PSNR} $\uparrow$ & \textbf{SSIM} $\uparrow$ & \textbf{LPIPS} $\downarrow$ \\ \hline
        E-DSSR~\cite{long2021dssr} & 13.398 $\pm$ 1.387 & 0.630 $\pm$ 0.057 & 0.423 $\pm$ 0.047 \\ \hline
        Ours w/o D & 24.088 $\pm$ 2.567 & 0.849$\pm$0.023  & 0.230 $\pm$ 0.023 \\
        Ours & \textbf{29.831 $\pm$ 2.208} & \textbf{0.925 $\pm$ 0.020} & \textbf{0.081 $\pm$ 0.022} \\ \bottomrule[0.5pt]
    \end{tabular}
    \label{tab:quanti_eval}
\end{table}

We present a qualitative ablation study on our depth-related modules in Fig. \ref{fig:ablation_depth_modules}. Without depth-supervision loss, we observe that the pipeline is not capable of learning correct geometry from single-viewpoint input. Moreover, when depth refinement is disabled, abrupt artifacts occur on the reconstruction results due to corruption in stereo depth estimation. Our depth-cueing ray marching can further diminish artifacts on 3D structures, especially for boundary points.


\section{Conclusion}

This paper presents a novel neural rendering-based framework for dynamic surgical scene reconstruction from single-viewpoint binocular images, as well as addressing complex tissue deformation and tool occlusion. We adopt the cutting-edge dynamic neural radiance field method to represent surgical scenes. In addition, we propose mask-guided ray casting to handle tool occlusion and impose stereo depth prior upon the single-viewpoint situation. Our approach has achieved superior performance on various scenarios in robotic surgery data such as large elastic deformations and tissue cutting. We hope the emerging NeRF-based 3D reconstruction techniques could inspire new pathways for robotic surgery scene understanding, and empower various down-stream clinical-oriented tasks.
\\
\\
\textbf{Acknowledgements.} This work was supported in part by CUHK Shun Hing Institute of Advanced Engineering (project MMT-p5-20), in part by Shenzhen-HK Collaborative Development Zone, and in part by Multi-Scale Medical Robotics Centre InnoHK.

%
%
%
\bibliographystyle{splncs04}
\bibliography{references}

\end{document}